# 🌷🤖 Tulip Agent – Enabling LLM-Based Agents to Solve Tasks Using Large Tool Libraries

Felix Ocker  Daniel Tanneberg  Julian Eggert  Michael Gienger

Honda Research Institute Europe
Germany, Offenbach am Main, 63073
firstname.lastname@honda-ri.de

## Abstract

We introduce `tulip agent`, an architecture for autonomous LLM-based agents with Create, Read, Update, and Delete access to a tool library containing a potentially large number of tools. In contrast to state-of-the-art implementations, `tulip agent` does not encode the descriptions of all available tools in the system prompt, which counts against the model's context window, or embed the entire prompt for retrieving suitable tools. Instead, the `tulip agent` can recursively search for suitable tools in its extensible tool library, implemented exemplarily as a vector store. The `tulip agent` architecture significantly reduces inference costs, allows using even large tool libraries, and enables the agent to adapt and extend its set of tools. We evaluate the architecture with several ablation studies in a mathematics context and demonstrate its generalizability with an application to robotics. A reference implementation and the benchmark are available at [github.com/HRI-EU/tulip_agent](github.com/HRI-EU/tulip_agent).

***Keywords*** Autonomous agent · Large language model · Tool library

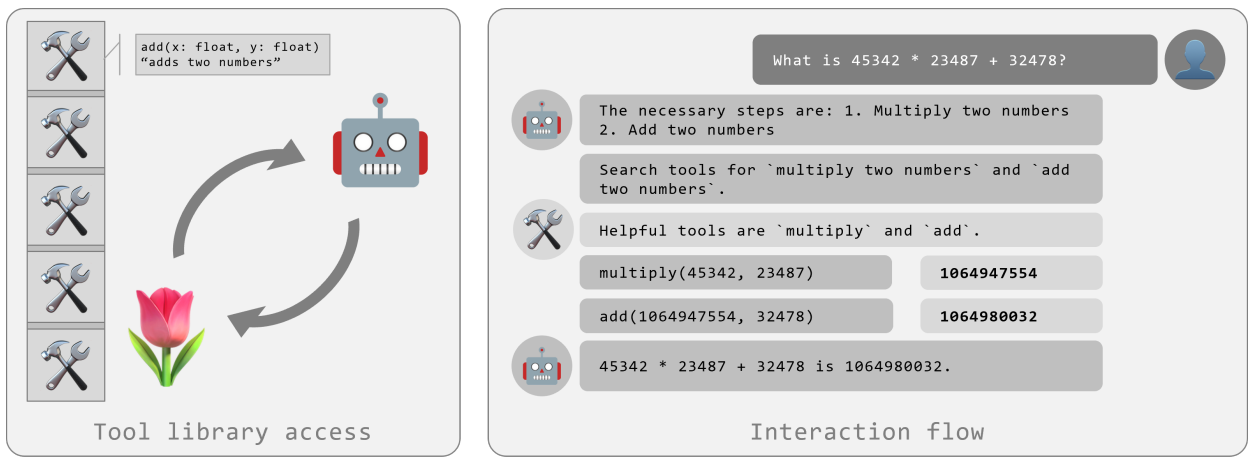

Figure 1: Exemplary application of the `tulip agent` architecture for solving a simple math problem.

# 1 Introduction

Advances in Large Language Models (LLMs) provide a technological basis for realizing autonomous agents. Examples range from assistants, e.g., pure software agents or on-device ones, to embodied AI in the form of robots. Key aspects for increasing autonomy in such agents are their understanding of their environment and their ability to act in it. This has become possible by LLMs being able to plan and to adhere strictly to instructions, also with regard to the format of their outputs, enabling them to use tools. Despite these advances and growing context windows for LLMs, there are still inherent limitations to current implementations resulting in the following Challenges 1 to 3:

**Challenge 1 (Costs)** *Tool descriptions count against the LLM's context window, driving costs both in terms of inference time and money.*

**Challenge 2 (Attention and tool limits)** *Choosing from a large number of tools is challenging for LLMs, as it imposes a form of the "needle-in-a-haystack" challenge[1]. This is due to LLMs struggling with in-context learning for long inputs Li et al. [2024], with retrieving multiple facts, and with having to reason about these facts [LangChain, 2024]. In addition, the number of tools that can be provided to the LLM may be limited, as is the case, e.g., for OpenAI models.*

**Challenge 3 (Staticity)** *Tool use is static and limited to a priori defined tools, limiting the adaptiveness of autonomous agents and their applicability to open-ended scenarios.*

This paper presents the `tulip agent` architecture to address these challenges.

# 2 Related Work

## 2.1 Autonomous Agents

The idea of autonomous agents has been around since decades. Such an agent can be "anything that can be viewed as perceiving its environment through sensors and acting upon that environment through actuators" Russell and Norvig [2016]. This definition can be extended to the include the "pursuit of [the agent's] own agenda [...] so as to effect what it senses in the future" Franklin and Graesser [1996].

Recent technological advances in LLMs have sparked an entire plethora of approaches for autonomous agents, including AutoGPT[2], Babyagi[3], and AgentGPT[4]. Similarly, Ge et al. [2023] present an LLM-based ecosystem, where agents have access to various applications.

Such LLM-based agents can operate in a loop consisting of receiving or retrieving inputs, planning, execution, and optionally reflecting on the results. Notably, providing relevant context increases the reasoning quality of LLMs [Wei et al., 2022], and they can benefit from agentic design patterns such as reflection, tool use, planning, and multi-agent collaboration [Ng, 2024]. For planning, specifically, LLMs benefit greatly from structured approaches such as Chain-of-Thought Wei et al. [2022], Tree-of-Thoughts Yao et al. [2023], and Monte Carlo Tree Search Zhang et al. [2024].

## 2.2 Tool Use for LLM-Based Agents

Qin et al. [2023a] generically define tool use as a three-step process. First, the task is decomposed, followed by reasoning for creating a plan and possibly adjusting the plan based on feedback from the environment, and finally solving each subtask by selecting appropriate tools. They recommend to train models specifically for generalized tool use and emphasize the challenge in trustworthy tool use and tool creation.

There are various efforts for creating LLMs specifically for using tools. Schick et al. [2023] present Toolformer, a model trained to call APIs. The model decides when and how to use tools in a self-supervised way and passes appropriate arguments. The authors showed improved performance on downstream tasks, but the model is limited to tasks that can be solved with a single tool call. Lu et al. [2023] leverage LLMs for multimodal question answering. They compose heterogeneous tools such as other LLMs, vision models, web search, and Python functions. An LLM-based planner is prompted with tool descriptions, decomposes the task decomposition, and infers a sequence of tool calls to execute based on in-context examples. For evaluating LLMs regarding their tool use capabilities, Huang et al. [2024] present a benchmark that involves six steps, namely planning, tool creation awareness, tool creation, tool usage awareness, tool selection, and tool usage. However, the tools are skeletons with descriptions but without executable implementations, and tool set sizes are limited to eight tools. Huang et al. [2024] showed that, in general, larger models perform better in tool utilization. They demonstrate the need for research regarding tool selection and creation, but do consider neither choosing tools from very large sets nor tool execution. There are also approaches for training smaller models with generalized tool-use capabilities. [Tang et al., 2023] present 7B and 13B parameter Alpaca derivatives finetuned on a tool-use corpus consisting of generated OpenAPI specifications and descriptions. The authors use ReAct Yao et al. [2022] for resolving user instructions, allowing multi-turn interactions. With a focus on on-device deployment, e.g., for Android devices, Chen and Li [2024] released a 2B parameter LLM trained for tool use. For efficiency, they use functional tokens for fine-tuning the model. This provides performance benefits but reduces generalizability compared to the Retrieval Augmented Generation (RAG)-based function calling mechanism in octopus-v1 Chen et al. [2024]. Chen et al. [2024] use Meta's FAISS for semantic search among functions, by embedding the user query and using the top five functions found. The model is evaluated with up to 20 functions in scenarios that require calling a single function, showing a reduction in context length and increased performance, but requiring function-set-specific finetuning. As of this writing, tool use is also supported by commercial models, e.g., the ones provided by OpenAI[5], Anthropic[6], and Cohere [7], as well as open-weights models such as Meta's Llama 3.1 [Llama Team, 2024]. However, even for these models, the number of tools that can be passed to the model is limited, for instance to 128 in the case of OpenAI.

---

[1] https://github.com/gkamradt/LLMTest_NeedleInAHaystack
[2] https://github.com/Significant-Gravitas/AutoGPT
[3] https://github.com/yoheinakajima/babyagi
[4] https://github.com/reworkd/AgentGPT

In contrast to approaches that focus on planning by itself, approaches such as ReAct Yao et al. [2022] and Reflexion Shinn et al. [2023] directly integrate tools and leverage feedback for plan adaptation. Following this scheme, Zhuang et al. [2023] present toolchain*, an approach using A* search for action space navigation. They focus on finding a cost-efficient solution path, representing the action space as a decision tree, where each node corresponds to an API call. Liu et al. [2023a] present an approach that relies on a task decomposer and a graph representation of tools and their dependencies in terms of inputs and outputs. During task decomposition, the user request is represented as a set of high-level subtasks, available inputs, and a formulation of the desired result. Based on these inputs, the graph is traversed using Depth First Search (DFS) and the tools are executed using an execution engine. This work focuses on multi-modal tools where input and output types are reliable indicators for choosing tools in combination with an LLM-based tool assessment for tool selection. Sun et al. [2023] present an approach for planning with tools including task decomposition and feedback from the environment. For an application to robotics, the authors rely on assertions to ensure that the environment is as expected. The authors distinguish open-loop planning with task decomposition from closed-loop planning with feedback. Further, they emphasize the difference of in-plan refinement, which adapts actions, and out-of-plan refinement, which adapts the plan itself.

Another relevant direction of research is the creation of tools by LLM-based agents themselves. Program-aided language models revolve around LLMs with code execution [Gao et al., 2023a]. Qian et al. [2023] present an approach for creating and using tools on the fly, and Cai et al. [2024] suggest combining a dedicated tool maker with a tool user. While versatile, these approaches do not focus on reusing the tools created and there is no easy way to integrate large sets of existing tools.

## 2.3  Retrieving Relevant Tools

Despite recent advances regarding increasing context windows, information retrieval from long-context windows is still problematic Li et al. [2024]. The performance of LLMs in this regard can, e.g., be measured with the needle-in-a-haystack test[8]. This test assesses performance on long contexts by inserting facts at various positions in the context window and rating whether the LLM is able to answer questions about these facts truthfully. Finding multiple needles has been shown to be even more challenging [LangChain, 2024], with performance degrading when the LLM has to retrieve more facts, or when the LLM has to reason about the facts retrieved.

---

[5] https://platform.openai.com/docs/guides/function-calling
[6] https://docs.anthropic.com/en/docs/build-with-claude/tool-use
[7] https://docs.cohere.com/docs/tool-use
[8] https://github.com/gkamradt/LLMTest_NeedleInAHaystack

One way to work around such limitations is via RAG. Lewis et al. [2020] define RAG as the combination of pre-trained parametric and non-parametric memory to augment LLMs by providing relevant background information. An LLM can serve as parametric memory while the non-parametric memory can be realized in various ways, including full-text search, search via sparse embeddings, such as BM25, search via dense embeddings, e.g., stored in a vector store, or query-based search for formalized databases [Gao et al., 2023b]. In cases of unstructured inputs, which do not necessarily use the same vocabulary as the non-parametric memory, semantic search is especially promising. Recent developments with regards to search, e.g., HNSW [Malkov and Yashunin, 2018], have enabled the development of performant vector stores.

This paradigm can be applied to tool search, providing additional context to the LLM in the form of available tools. For instance, Qin et al. [2023b] finetuned Llama for tool use in the form of 16.000 REST APIs, resulting in ToolLLaMA. The authors trained an API retriever that searches among the API descriptions based on the complete task without decomposition. This is combined with multi-round decision making of the LLM in the form of DFS-based tool selection and execution for single-tool and multi-tool scenarios. Patil et al. [2023] present Gorilla, a LLaMA-based model finetuned for writing API calls. The authors compare a zero-shot approach with a retriever-based approach (BM25 or dense embeddings) with `top_k = 1` and an oracle retriever. The execution of API calls is not part of the evaluation, and the tasks are limited to using a single tool and single-step reasoning. A key learning is that a retriever may be beneficial, but adding a non-optimal retriever may misguide the model and result in more errors. Another work suggests to clearly separate tool selection and tool usage [Anonymous, 2024]. Here, first a description for the tool needed is generated from the user query based on which the retriever identifies the five most relevant tools. Then, an LLM is employed to actively analyze the `top_k` tools found in a separate step. Finally, the tool call is executed. In this approach the LLM cannot actively trigger searches for tools, it is limited to tasks that require exactly one tool, and the approach also increases overall costs.

## 2.4 Application of LLMs with Tool Access to Robotics

There is a big interest in utilizing the remarkable commonsense reasoning abilities of LLMs in robotic systems. LLMs have been leveraged for robotic applications in different ways and for different purposes [Kira, 2022, Zeng et al., 2023], from generating high-level robotic plans [Joublin et al., 2024, Liu et al., 2023b, Zhou et al., 2023, Huang et al., 2022], to generating code for controlling the robot [Vemprala et al., 2023, Liang et al., 2023, Wu et al., 2023, Singh et al., 2023], to steering the robot's behavior in human-robot interaction setups [Tanneberg et al., 2024, Wang et al., 2024]. When used for planning, typically the available skills, like motion primitives or simple manipulation actions, are given and the size of this library is limited [Ahn et al., 2022, Lin et al., 2023, Hazra et al., 2024]. The LLMs are used to decompose higher level tasks into sequences of these skills, often utilizing formal representations, to solve the task [Liu et al., 2023c, Silver et al., 2024]. For intelligent robots that grow their knowledge and skills in an open-ended way [Yu et al., 2023, Xie et al., 2024], these skill libraries grow [Zhang et al., 2023, Zhou et al., 2023] and it becomes harder for the LLM to identify the useful skills for a given task. Hence, besides lower costs, the `tulip agent` offers a framework to deal with large and continuously growing skill libraries.

## 3 Tulip Agent Architecture

The `tulip agent` architecture aims to address Challenges 1 to 3 by providing an LLM with access to an extensible tool library. This enables access to an arbitrarily large set of tools that can be efficiently extended and adapted while reducing overall costs.

Figure 2 gives an overview of `tulip agent`'s components, with the arrows indicating the flow of information. To set up a `tulip agent`, information about the available tools, in our case Python functions, is extracted automatically via code introspection. From the functions' docstrings, we generate embeddings which are stored in the tool library together with an LLM-compatible representation of the tool information ❶. When receiving a user prompt ❶, the model decomposes the request into subtasks and passes respective descriptions to the search module ❷. These descriptions are the basis for embeddings, which the search module uses to find suitable tools for each subtask via semantic search. The search module passes the information about the most relevant tools to the model ❸, which calls appropriate tools for all subtasks ❹. The tools are executed accordingly ❺ and the results are fed back to the model ❻, allowing the model to initiate further actions or provide a response to the user ❼.

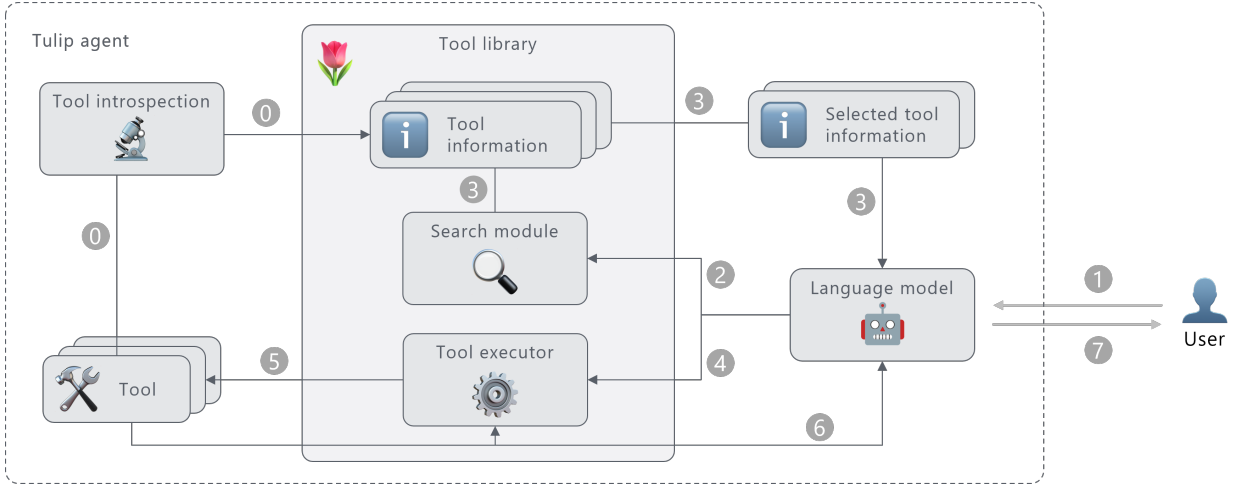

Figure 2: Overview of the `tulip agent` architecture and its information flow.

The following describes the `tulip agent` architecture in its default setup with Chain of Thought (CoT) prompting for step-by-step task decomposition and tool execution, the `CotTulipAgent`. Several variations and their differences compared to this default setup are described in Section 3.6.

## 3.1 Problem Formulation

We propose the `tulip agent` architecture for LLM-backed autonomous agents. Such an agent is initialized with a potentially large set of tools $\mathcal{T}$, for which natural language descriptions and documentation are available. When prompted with a natural language query $q \in \mathcal{Q}$ from the task space $\mathcal{Q}$, the agent's task decomposition model $\mathcal{M}_{td}$ must first decompose the query into a plan $\mathcal{P}$ consisting of a sequence of subtasks such that the subtasks can be resolved with the tools available. Optionally, the model $\mathcal{M}_{td}$ can be primed with selected information $\mathcal{I}_\mathcal{T}$ about the tools available. In Figure 2, we refer to this step as *task decomposition* as performed between ❶ and ❷.

$$\mathcal{M}_{td}(q, \mathcal{I}_\mathcal{T}) \to \mathcal{P} \qquad (1)$$

Second, the selection model $\mathcal{M}_s$ has to select suitable tools $\mathcal{T}_i^*$ for each subtask $i$ from the potentially large set of tools $\mathcal{T}$. In particular, this step may comprise *tool retrieval*, see ❷→❸ in Figure 2. This can be formulated as a search problem, optionally with a search tool $t_s$.

$$\mathcal{M}_s(\mathcal{P}, \mathcal{T}, t_s) \to \{\mathcal{T}_i^*\}_i \qquad (2)$$

Third, the extraction model $\mathcal{M}_e$ must *extract relevant input values* $V_{in}$, i.e., the parameters for the tools. These may be available from the user's query $q$, as information available from prior tool use, i.e., previous output values $V_{out}^-$, or as the model's common-sense knowledge, for every subtask. In Figure 2, this is done before step ❹.

$$\mathcal{M}_e(\mathcal{T}_i^*, q, V_{out}^-) \to V_{in} \qquad (3)$$

We refer to the combination of a selected tool and the parameters required as an action $a_i \in \mathcal{A}$ from the action space $\mathcal{A}$. These actions are composed by the model $\mathcal{M}_a$. In Figure 2, such actions are passed from the model to the tool executor as *tool calls* in step ❹.

$$\mathcal{M}_a(\mathcal{T}_i^*, V_{in}) \to a_i \qquad (4)$$

Execution of the actions $a_i \in \mathcal{A}$ by the tool executor $\mathcal{E}$ results in output values $V_{out}$. These output values can be new information to be processed further, failure feedback, or the final response to the user query.

$$\mathcal{E}(a_i) \to V_{out} \qquad (5)$$

Using this output, the model can generate adapted actions, until the plan $\mathcal{P}$ is completed and the agent can return a final response to the user. This *execution of actions* corresponds to ❺→❼ in Figure 2.

Note that the various models $\mathcal{M}_x$ may be implemented as a single language model prompted with different system prompts.

### 3.2 The Tools and the Tool Library

In the context of this paper, a tool is an executable function that fulfils a purpose and returns either a result or a status message. Examples include mathematical functions, as provided by a calculator, but also calls to a robot's API. To facilitate providing tools to the `tulip agent`, entire files with functions are imported.

For using tools, LLMs require a unique identifier for the tool, which can be resolved to call the tool, a description of the tool's purpose, and names, types, and descriptions of necessary input parameters. Our approach relies on a function analyzer for introspection. It can ingest Python functions documented according to the Sphinx style[9], but this could be extended for other types of tools. From the information extracted, we construct a tool library. In principle, the tool library may be any kind of database that supports searching for appropriate tools. Since the user provides natural language inputs and LLMs work well for natural language, semantic search via dense embeddings that allow matching the task to available tools is especially promising. During initialization of the agent, we create embeddings of the function names and the corresponding docstrings, i.e., vector representations capturing the semantics, and store them together with the function descriptions generated by the introspection module. The resulting vector store allows searching for suitable tools via semantic search. The process for initializing the tool library ❶ is summarized in Algorithm 1.

---

**Algorithm 1** Initializing the tool library.

1: **procedure** INITIALIZETOOLLIBRARY(*modulenames*, *directory*)
2:    $VectorStore \leftarrow$ INITIALIZEVECTORSTORE(*directory*)         ▷ Initialize an empty vector store
3:    **for** *modulename* ∈ *modulenames* **do**
4:       *module* ← IMPORT(*modulename*)
5:       **for** *function* ∈ *module.functions* **do**
6:          *data* ← ANALYZE(*function*)   ▷ Get function name, docstring, and parameter names, descriptions, and types
7:          *embedding* ← EMBED(*data*)
8:          $VectorStore.add(id : data.name, document : data, embedding : embedding)$
9:          $ToolLookup[data.name] \leftarrow function$

---

### 3.3 Task Decomposition and Tool Retrieval

Upon initialization, the `tulip agent` may take natural language user inputs describing the task to be fulfilled. Compared to the granularity of available tools, such tasks are typically posed on a high level of abstraction. Thus, they cannot be matched sensibly to individual tools. To cope, we use an LLM to decompose the task into subtasks, see Listing 2 for the `CotTulipAgent`'s system prompt, and Listing 3 for splitting the task into steps. Granularity and clarity are essential to reduce the semantic distance between the task description and the descriptions of suitable tools. This is because unnecessary specifics in the task description would add noise and might result in more ambiguous tool suggestions. During decomposition, we have the LLM create more generic descriptions for the subtasks, which can be matched better to the generic descriptions of tools. This results in the plan $\mathcal{P}$ in the form of a list of generic natural language descriptions of subtasks.

The `tulip agent` searches its tool library for appropriate tools for each subtask. Specifically, an embedding is created for each subtask in the plan, which is matched against the tool descriptions' embeddings, returning the `top_k` most suitable tools.

Notably, the `tulip agent` architecture supports a recursive decomposition and search for tools. This is relevant in case the initial subtasks are not fine-grained enough to find suitable tools. By setting a similarity threshold for the semantic search, we can ensure that only suitable tools are returned. In case no tools are found whose descriptions are sufficiently similar to the task description the agent decomposes the subtask further, conducting another search for the next level of tools. To avoid infinite loops, we resort to setting a maximum recursion depth.

---

[9]https://sphinx-rtd-tutorial.readthedocs.io/en/latest/docstrings.html

The search for tools in the tool library as a semantic search is summarized in Algorithm 2. By default, we use the squared l2 norm as a distance function between embedding vectors.

---

**Algorithm 2** Searching for tools in the tool library.
---
1: **procedure** SEARCHTOOLS(*tasks*)
2:   *tools_by_task* ← ∅
3:   **for** *task* ∈ *tasks* **do**
4:     *embedding* ← EMBED(*task*)
5:     *tools* ← TOOLLIBRARY.SEMANTICSEARCH(*task*, *top_k*, *similarity_threshold*)
6:     **if** *tools* = ∅ **then**
7:       *decomposition_prompt* ← FORMAT(*base_decomposition_prompt*, *task*)   ▷ Cp. Listing 3
8:       *subtasks* ← QUERYLLM(*decomposition_prompt*)
9:       *tool_search_prompt* ← FORMAT(*base_tool_search_prompt*, *subtasks*)   ▷ Cp. Listing 4
10:       *tool_search_strings* ← QUERYLLM(*tool_search_prompt*)
11:       *subtools_by_subtask* ← SEARCHTOOLS(*tool_search_strings*)
12:       *tools_by_task*.extend(*subtools_by_subtask*)
13:     **else**
14:       *tools_by_task*.append((*task*, *tools*))
15:   **return** *tools_by_task*

---

### 3.4 Tool Use

Based on the plan $\mathcal{P}$ consisting of subtasks and the identified tools, the LLM is prompted to generate tool calls, see Listing 5. This is done in a step-by-step way, allowing the LLM to take into account previous return values. Note that tool calls can take the form of structured JSON responses, as is the case for models fine-tuned for tool use, or text that has to be parsed. For each tool call, the respective tool is retrieved from the lookup, see *ToolLookup* in Algorithm 1, which is then executed by the tool executor $\mathcal{E}$ with the parameters provided. The tool's return value can be an intermediate result, e.g., from a calculation, a status message, e.g., from a robot action, or feedback about a failure. This result is fed back to the LLM, enabling the model to react by calling further tools or even searching for other relevant tools in the tool library. When all subtasks are solved or if it becomes clear that the LLM is unable to solve the task, the LLM may provide a final response to the user.

The query process for the `CotTulipAgent` is summarized in Algorithm 3. It comprises several steps, first splitting the task imposed by the user into subtasks, searching for suitable tools for each subtask, using these tools, and responding with the final result.

---

**Algorithm 3** `CotTulipAgent` query process.
---
1: **procedure** QUERY(*user_query*)
2:   *decomposition_prompt* ← FORMAT(*base_decomposition_prompt*, *user_query*)   ▷ Cp. Listing 3
3:   *subtasks* ← QUERYLLM(*decomposition_prompt*)
4:   *tool_search_prompt* ← FORMAT(*base_tool_search_prompt*, *tasks*)   ▷ Cp. Listing 4
5:   *tool_search_strings* ← QUERYLLM(*tool_search_prompt*)
6:   *tools* ← SEARCHTOOLS(*tool_search_strings*)
7:   *execution_prompt* ← FORMAT(*tool_prompt*, *subtasks*)   ▷ Cp. Listing 5
8:   *response* ← QUERYLLM(*execution_prompt*, *tools*)
9:   **while** *response.tool_calls* **do**
10:     *intermediate_results* ← ∅
11:     **for** *tool_call* ∈ *response.tool_calls* **do**
12:       *tool_call_data* ← RESOLVE(*tool_call*)
13:       *intermediate_result* ← TOOLLIBRARY.EXECUTE(*tool_call_data.name*, *tool_call_data.parameters*)
14:       *intermediate_results*.append(*intermediate_result*)
15:     *response* ← QUERYLLM(*execution_prompt*, *tools*, *intermediate_results*)
16:   **return** *response*

---

## 3.5 Autonomous Tool Management

To further increase the versatility of the agent and enable it to learn in a continuous way, see Challenge 3, we provide the `tulip agent` variant `AutoTulipAgent` with five generic tools. These tools enable it to plan and provide it with Create, Read, Update, and Delete (CRUD) access to its tool library. The first two tools allow the agent to decompose tasks and search for tools in the tool library, as described in Section 3.3. In contrast to other variants, the `AutoTulipAgent` does not follow a predefined flow, though, but may call these tools as necessary, potentially recursively. In addition, the agent disposes of tools for creating new tools, updating existing tools, and deleting existing ones. These tools allow the agent to change its own tool library on-the-fly. While tool deletion is trivial, creating new tools and updating existing ones are similar to each other in their approach. Leveraging the code generation abilities of LLMs, the agent triggers the generation of a new tool for a specific task described in natural language. When updating existing tools, the existing code is fed into the code generation LLM as additional context. To ensure validity of the new tool, the agent checks its executability in a loop in the spirit of Joublin et al. [2024] before saving. For Python tools, this can be achieved, e.g., via the `ast` module. Note that the LLM must generate code that can be processed by the `tulip agent`'s function analyzer. Specifically, the code must be properly documented using descriptions and type hints. The resulting code is written to a file, and automatically introspected, embedded, and loaded into the tool library. Using an importer, specifically Python's `importlib`, the new tool can be loaded dynamically, making it callable during operation. Eventually, the `AutoTulipAgent` may use the extended tool library to solve the initial problem.

The benefit of the `AutoTulipAgent` over a code interpreter lies in its ability to load large sets of available tools, reusability of tools leading to reduced costs, and the ability to continuously improve existing tools.

## 3.6 Architecture Variations

The `tulip agent` architecture can be used in conjunction with other LLM agent paradigms. Using a *tool library* is characteristic for the `tulip agent` architecture, cp. Section 3.2. This also implies *tool use*, i.e., the LLM can initiate tool calls, which are resolved and the results of which are provided to the LLM. To further improve the LLM's performance, CoT prompting can be applied to guide the LLM through the process in a fixed way, independent of the other features. In addition, we investigate several variations for priming the LLM used for task decomposition towards the tools available in the tool library. Finally, we investigate equipping the agent with means for *self-editing*, specifically editing its own tools. In the case of the `tulip agent`, this is achieved via CRUD operations for the tool library available to the agent, cp. Section 3.5. Table 1 gives an overview of all the agent variations used, including baselines without access to a tool library and even without tool access altogether.

Table 1: Agent variations and their features (∘ baseline setups, ◁ ablations, ◇ main method, ○ variation, ▷ experimental). CRUD refers to creating, reading, updating, and deleting tools in the tool library.

|                              | Tools | CoT | **Tool library** | CRUD | Comment                                 |
|------------------------------|-------|-----|------------------|------|-----------------------------------------|
| `BaseAgent`∘                 |       |     |                  |      | Off-the shelf LLM                       |
| `NaiveToolAgent`∘            | x     |     |                  |      | LLM with tool access                    |
| `CotToolAgent`∘              | x     | x   |                  |      | CoT prompting                           |
| `MinimalTulipAgent`◁         | x     |     | x                | R    | Tool search based on user prompt        |
| `NaiveTulipAgent`◁           | x     |     | x                | R    | Agent searches for tools                |
| `CotTulipAgent`◇ (CTA)       | x     | x   | x                | R    | `CotToolAgent` ∪ `NaiveTulipAgent`      |
| `InformedCotTulipAgent`○     | x     | x   | x                | R    | `CTA` with brief library description    |
| `PrimedCotTulipAgent`○       | x     | x   | x                | R    | `CTA` with selected tool names          |
| `OneShotCotTulipAgent`○      | x     | x   | x                | R    | `CTA` with example                      |
| `AutoTulipAgent`▷            | x     |     | x                | CRUD | Autonomous agent                        |

Note that the number `top_k` of tools to be returned may depend on the agent type. For instance, the `MinimalTulipAgent` requires more tool options to choose from as it does not split the user prompt into more specific tasks.

# 4 Experiment Setup and Results

## 4.1 Implementation

The prototype is implemented in Python 3.10 and the tool library is built on ChromaDB[10]. As language models, we used OpenAI models via the API, specifically `gpt-4-turbo-2024-04-09` and `gpt-3.5-turbo-0125`, and `text-embedding-ada-002`, `text-embedding-3-small`, and `text-embedding-3-large` as embedding models. To achieve results that are as deterministic as possible, we set the temperature to $1e-9$ and repeated experiments five times. Note that we report the interquartile mean for costs. This is because the API interaction limit of 100 may lead to outliers that would skew the comparison.

## 4.2 Quantitative Evaluation

We evaluate the `tulip agent` architecture, several ablations, and baselines regarding correctness and costs. Correctness includes high-level correctness of the result, independently of how it is reached, and precision and recall with regard to the tools used. Costs are based on token counts, including input tokens, output tokens, and embedding tokens. The monetary assessment of costs considers current OpenAI pricing, see Table 5. Note that to avoid bias in favor of the `tulip agent` architecture, the costs for these variations always include the embedding costs for setting up the entire tool library. In practice, the costs for creating the tool library are only incurred once, further reducing overall costs when used multiple times.

We designed a benchmark with three categories with 20 tasks each.

- Easy: Requires using one tool
- Medium: Requires using two or three tools
- Hard: Requires using at least four tools

Note that this classification imposes a harder assessment compared to Liu et al. [2023a]. For medium and hard tasks, the task has to be split up into subtasks, requiring planning. For an example of a hard task including the ground truth with regards to expected results and expexted functions to be used see Listing 10.

To set up the tool library, we generated 100 simple math functions with type hints and docstrings in Sphinx format using an LLM, and validated them manually. For details regarding their creation see Section 5. This set of tools includes examples such as `add()`, `multiply()`, and `coefficient_of_variation()`, see Listings 6 to 8, respectively. We limited the number of tools in the experiments to 100 to enable comparisons with `CotToolAgent`, which is limited by OpenAI's current tool limit of 128. However, we qualitatively demonstrated the `tulip agent` architecture to work also for a significantly larger tool set containing 600 tools.

The results for various agents are shown in Figure 3. Note that all variations introduced in Section 3.6 are included except for `AutoTulipAgent`, as this one does not work reliably enough with less performant models such as `gpt-3.5-turbo-0125`.

This comparison of variants led to three findings. First, we learned that tools are essential for current LLMs to solve complex tasks reliably, cp. Finding 1. This insight becomes apparent from the comparison of agent variations with tool use and the `BaseAgent` regarding correct results. This argument also aligns with the idea of using specialized agents in a Multi Agent System (MAS), only that in our case an agent may also be part of a tool, as is the case for task decomposition.

**Finding 1 (Tools are essential for solving complex tasks)** *Using tools allows LLMs to reliably solve complex tasks such as in the mathematics benchmark, similar to how a human would use a calculator. In addition, tools are of paramount importance for enabling an agent to interact with its environment beyond pure text generation. This is crucial for applications such as robotics.*

Second, we found that using a tool library to give an agent access to tools reduces costs significantly in most scenarios, cp. Finding 2. This connection is emphasized with increasing numbers of tools or when solving more complex tasks, which require several calls to the LLM. Note that a generic issue for costs lies in the possiblity of the LLM getting stuck in a loop of responses.

---

[10]https://www.trychroma.com/

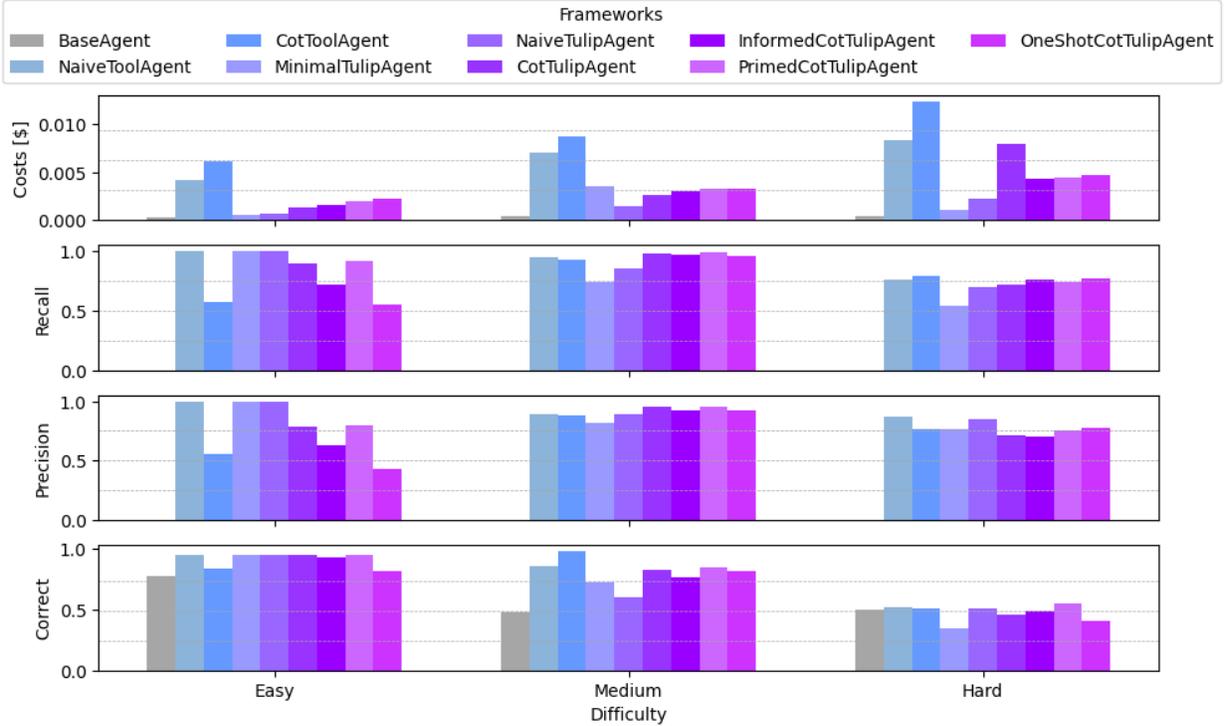

Figure 3: Performance of several agents regarding costs in USD, tool use precision, tool use recall, and correctness of the result on several math tasks. Settings: language model `gpt-3.5-turbo-0125`, embedding model `text-embedding-3-large`, `top_k = 5`, 5 runs.

**Finding 2 (Using a tool library significantly reduces costs)** *In our evaluation, the `tulip agent` architecture significantly reduces costs, and variants such as `CotTulipAgent` or `PrimedCotTulipAgent` maintain correctness. This is because costs for embeddings are negligible, the number of input tokens is drastically reduced, and there are only relatively few output tokens necessary. For our experiments, we found the cost reduction to be a factor of two to three when using 100 tools without excessively long documentation. Factors that further increase the benefits of the `tulip agent` are large sets of functions and complex tasks that require calling many tools, as all the tools are passed to the LLM for each subtask. Conversely, in scenarios that include only very simple tasks and few tools, using the `tulip agent` architecture may not yield benefits.*

Third, planning, i.e., task decomposition, positively influences tool use and thus overall performance, cp. Finding 3. This step can be improved even further by priming the LLM used for task decomposition with an overview of the available tools. This finding is in line with related work [Patil et al., 2023].

**Finding 3 (Task decomposition improves tool use)** *Decomposing tasks into subtasks that can be executed with individual tools significantly improves the results, see `MinimalTulipAgent` versus `CotTulipAgent`. Priming the task decomposition with available tools further improves performance, see `CotTulipAgent` versus `PrimedCotTulipAgent`.*

To assess the impact of the LLM used within the agent, we ran several experiments with various agents, including the baselines `BaseAgent` and `CotToolAgent`. The results are summarized in Table 2. Unsurprisingly, `gpt-4-turbo-2024-04-09` performs best, but is also the most expensive. These experiments confirmed that the `tulip agent` variants lead to a reduction in costs while providing the same level of correct task completion, cp. Finding 2, even across models. However, agents with advanced functionality, specifically CRUD access to the tool library, require better LLMs, cp Finding 4.

**Finding 4 (Language model performance influences the suitability of agent designs)** *With greater functionality, the agent variations require a more performant LLM. For instance, `AutoTulipAgent` re-*

Table 2: Language model impact: Performance of several agents regarding costs per task in USD ($) and correctness of the result (%) on several math tasks. Row-wise best variants are marked bold, column-wise best combinations w.r.t. correctness are underlined. We conducted 5 runs with $top\_k = 5$ and `text-embedding-3-large` as embedding model.

| LLM | | BaseAgent % | BaseAgent $ | CotToolAgent % | CotToolAgent $ | CotTulipAgent % | CotTulipAgent $ | PrimedCotTulipAgent % | PrimedCotTulipAgent $ |
|---|---|---|---|---|---|---|---|---|---|
| gpt-3.5-turbo-0125 | Easy | 0.78 | **0.000** | 0.84 | 0.006 | **0.95** | 0.001 | **0.95** | 0.002 |
| | Medium | 0.48 | **0.000** | **0.98** | 0.009 | 0.83 | 0.003 | 0.85 | 0.003 |
| | Hard | 0.50 | **0.001** | 0.51 | 0.012 | 0.46 | 0.008 | **0.55** | 0.004 |
| gpt-4-turbo-2024-04-09 | Easy | <u>0.98</u> | **0.006** | 0.97 | 0.099 | **<u>1.00</u>** | 0.024 | **<u>1.00</u>** | 0.037 |
| | Medium | <u>0.64</u> | **0.010** | **<u>1.00</u>** | 0.173 | <u>0.99</u> | 0.049 | <u>0.90</u> | 0.066 |
| | Hard | <u>0.74</u> | **0.013** | **<u>1.00</u>** | 0.250 | <u>0.96</u> | 0.079 | 0.85 | 0.089 |
| gpt-4o-2024-05-13 | Easy | 0.97 | **0.003** | **<u>1.00</u>** | 0.069 | **<u>1.00</u>** | 0.018 | **<u>1.00</u>** | 0.020 |
| | Medium | 0.53 | **0.006** | **0.86** | 0.084 | 0.83 | 0.027 | 0.83 | 0.037 |
| | Hard | 0.70 | **0.008** | **0.97** | 0.121 | <u>0.96</u> | 0.044 | <u>0.94</u> | 0.056 |

*quires models on the level of `gpt-4-turbo` with regards to instruction following capabilities. Otherwise issues may arise with generated tools, for example with invalid documentation.*

To assess the influence of the choice of the embedding model on the `tulip agent` variants' performance, we conducted experiments with `text-embedding-ada-002`, `text-embedding-3-small`, and `text-embedding-3-large`. Table 3 summarizes the results for `MinimalTulipAgent`, `CotTulipAgent`, `InformedCotTulipAgent`, and `PrimedCotTulipAgent`. The baseline models are not included in the comparison as they are not affected by the choice of the embedding model.

Table 3: Embedding model impact: Performance of several agents regarding costs per task in USD ($) and correctness of the result (%) on several math tasks. Row-wise best variants are marked bold, column-wise best combinations w.r.t. correctness are underlined. We conducted 5 runs with $top\_k = 5$ and `gpt-3.5-turbo-0125` as language model.

| Embedding model | | MinimalTulipAgent % | MinimalTulipAgent $ | CotTulipAgent % | CotTulipAgent $ | InformedCotTulipAgent % | InformedCotTulipAgent $ | PrimedCotTulipAgent % | PrimedCotTulipAgent $ |
|---|---|---|---|---|---|---|---|---|---|
| text-embedding-ada-002 | Easy | <u>0.95</u> | **0.001** | <u>0.95</u> | 0.001 | <u>0.95</u> | 0.002 | **0.96** | 0.002 |
| | Medium | 0.61 | 0.004 | 0.82 | **0.003** | 0.81 | 0.003 | **0.84** | 0.003 |
| | Hard | 0.27 | **0.002** | 0.49 | 0.004 | <u>0.53</u> | 0.004 | **0.55** | 0.005 |
| text-embedding-3-small | Easy | **<u>0.95</u>** | **0.000** | 0.93 | 0.001 | 0.91 | 0.002 | 0.94 | 0.002 |
| | Medium | 0.55 | **0.001** | <u>0.85</u> | 0.003 | <u>0.82</u> | 0.003 | **<u>0.87</u>** | 0.003 |
| | Hard | 0.30 | 0.004 | <u>0.50</u> | 0.005 | 0.52 | 0.009 | **<u>0.56</u>** | 0.005 |
| text-embedding-3-large | Easy | **0.95** | 0.001 | **0.95** | 0.001 | 0.93 | 0.002 | **0.95** | 0.002 |
| | Medium | <u>0.72</u> | 0.004 | 0.83 | **0.003** | 0.77 | 0.003 | 0.85 | 0.003 |
| | Hard | <u>0.35</u> | **0.001** | 0.46 | 0.008 | 0.49 | 0.004 | **0.55** | 0.004 |

While `MinimalTulipAgent` is the cheapest variant, it performs worst except for easy tasks. `PrimedCotTulipAgent` on the other hand performs best, which is consistent with Finding 3. Interestingly, this holds across embedding models, cp. Finding 5. This seems reasonable, as even the smallest model, `text-embedding-ada-002`, performs well in embedding benchmarks[11]. The only exception is `MinimalTulipAgent`, which performs well for easy tasks and shows substantial benefits from more capable embedding models. Note that the costs for generating embeddings is almost irrelevant, due to the significantly higher costs for output tokens.

**Finding 5 (Embedding model performance has little influence on tool retrieval)** *For variants with more advanced planning, the choice of the embedding model barely influenced the correctness of results. However, this may be due the already high level of capability of all of the models included.*

---
[11]`https://huggingface.co/spaces/mteb/leaderboard`

To assess the influence of the pre-selection of tools during semantic search compared to the selection by the LLM, we varied the `top_k` parameter for several `tulip agent` variations, see Table 4. In particular the comparison of `MinimalTulipAgent` with the other variants shows that sound task decomposition allows searching more narrowly, i.e., using lower values for `top_k`, cp. Finding 6. In the naive case of `MinimalTulipAgent`, this is because the embedding of the user query is directly compared to all tools and thus must be at least as high as the number of tools needed for successfully completing the task, resulting in a drastic increase in performance for hard tasks with higher `top_k`.

Table 4: `top_k` impact: Performance of several agents regarding costs per task in USD ($) and correctness of the result (%) on several math tasks. We conducted 3 runs with `gpt-3.5-turbo-0125` as language model and `text-embedding-3-large` as embedding model. Row-wise best variants are marked bold, column-wise best combinations w.r.t. correctness are underlined. We conducted 3 runs with `gpt-3.5-turbo-0125` as language model and `text-embedding-3-large` as embedding model.

| top_k tool retrieval | | Minimal TulipAgent % | $ | Cot TulipAgent % | $ | InformedCot TulipAgent % | $ | PrimedCot TulipAgent % | $ |
|---|---|---|---|---|---|---|---|---|---|
| $top\_k = 1$ | Easy | **0.95** | **0.000** | **0.95** | 0.001 | 0.92 | 0.001 | **0.95** | 0.002 |
|  | Medium | 0.38 | **0.001** | 0.83 | 0.002 | **0.88** | 0.002 | 0.80 | 0.002 |
|  | Hard | 0.28 | **0.001** | 0.45 | 0.002 | 0.47 | 0.003 | **0.53** | 0.003 |
| $top\_k = 5$ | Easy | **0.95** | 0.001 | **0.95** | 0.001 | 0.93 | 0.002 | **0.95** | 0.002 |
|  | Medium | 0.72 | 0.004 | 0.83 | **0.003** | 0.77 | 0.003 | **0.85** | 0.003 |
|  | Hard | 0.35 | **0.001** | 0.46 | 0.008 | 0.49 | 0.004 | **0.55** | 0.004 |
| $top\_k = 10$ | Easy | **0.95** | 0.001 | 0.93 | 0.002 | 0.87 | 0.002 | 0.90 | 0.008 |
|  | Medium | **0.83** | 0.001 | 0.80 | 0.004 | **0.83** | 0.004 | 0.82 | 0.004 |
|  | Hard | 0.48 | 0.002 | **0.57** | 0.012 | 0.52 | 0.005 | 0.53 | 0.005 |

**Finding 6 (Better planning allows narrower search)** *`tulip agent` variants with better planning, especially `PrimedCotTulipAgent`, benefit less from increasing top_k beyond 5. More naive implementations, such as `MinimalTulipAgent`, on the other hand, benefit from an increase in top_k, especially for hard tasks.*

Finding 6 hints towards the benefit of recursive task decomposition for more complex tasks. However, in our experiments, recursive task decomposition did not yield benefits, as a single-step decomposition turned out to be sufficient for these kinds of tasks.

### 4.3 Qualitative Demonstration of CRUD Operations

The `AutoTulipAgent` with CRUD operations can be applied in scenarios where suitable tools are not specified a priori. In such cases, the tool search does not return suitable tools, and the agent creates a new tool using an LLM, cp. Section 3.5. For instance, it can solve the following sequence of tasks starting with an empty tool library, cp. Listing 13.

1. What is the square root of 23456789?
2. Change the square root tool to correctly work for negative numbers.
3. Calculate the square root of -200.
4. Delete the square root tool.

This example demonstrates the self-editing capabilities of the `tulip agent` architecture, and anecdotal evidence suggests that this approach works well for sensibly encapsulated functions. Note that this approach greatly depends on the LLM's abilities to generate code and that it may be sensible to use a dedicated model for realizing the tool creation.

**Finding 7 (Autonomous creation of a tool library)** *The `tulip agent` architecture is suited for continually creating tools and building a tool library on the fly. This, however, depends largely on the abilities of the LLM used with regard to code generation.*

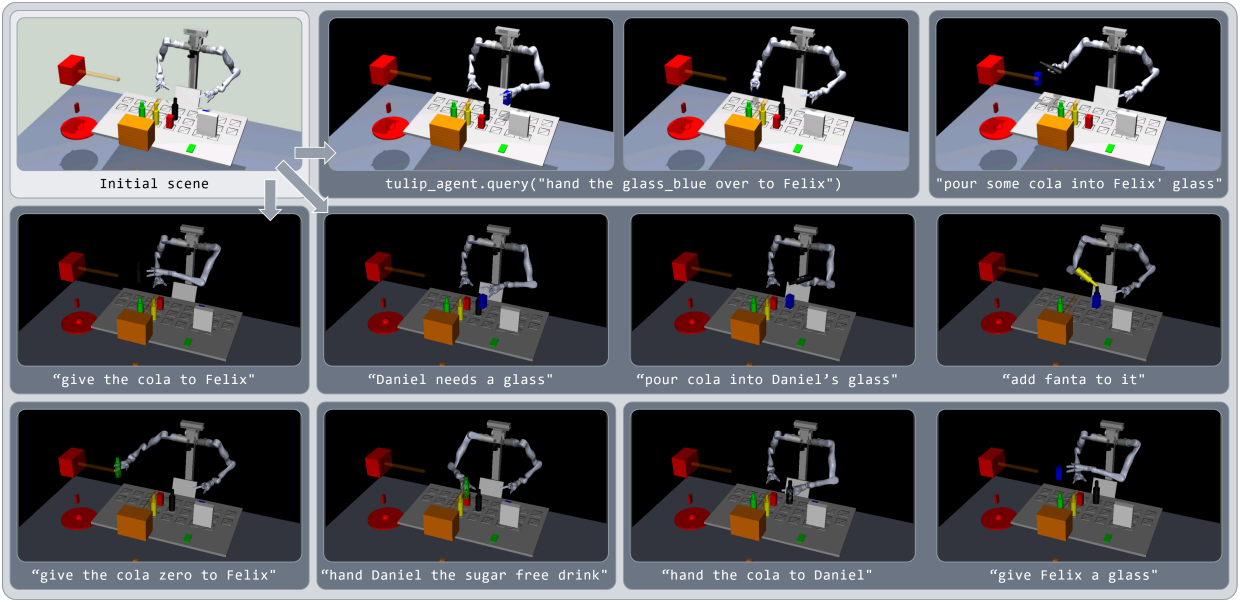

Figure 4: The `CotTulipAgent` controlling a supportive robot in a simulation, assisting two humans (*Felix* in *red*, *Daniel* in *orange*) with various tasks within a tabletop scenario with different drink options.

### 4.4 Application to Robotics

To assess the applicability of the `tulip agent` architecture to embodied agents, we also applied it to a robotics scenario. We built on top of a publicly available simulation of a proactive supportive robot [Tanneberg et al., 2024]. In its original form, the robot is controlled by an LLM with access to a variety of tools for information retrieval, physical actions, and social expressions. We initialized a `CotTulipAgent`-backed agent with access to the same tools and extended the agent's system prompt with the original robot character [Tanneberg et al., 2024]. This setup worked in a plug-and-play way and we tested it in various situations, see Figure 4 for some examples. The `CotTulipAgent`-backed agent successfully solves the given instructions utilizing its tool library consisting of actions such as `pour_into` and `hand_over`, cp. Listing 9.

While the number of tools is limited in comparison to the math setup, an extensible tool library has the potential to enable the robot to continuously learn and adapt in an open world [Zhang et al., 2023, Tziafas and Kasaei, 2024].

## 5 Summary and Outlook

This paper presented the `tulip agent` architecture. Using a dedicated tool library, it provides three benefits. First, it significantly reduces costs, both in terms of time and money, when using large numbers of tools (Challenge 1). Second, it allows working with large numbers of tools despite context window limitations and the needle-in-a-haystack challenge (Challenge 2). Third, it allows for dynamic creation and loading of tools (Challenge 3), paving the way towards continuously evolving autonomous agent systems. Based on several ablations regarding the LLM, the embedding model, the parameter `top_k` for tool search, and the `tulip agent` architecture, we report on various findings that are hopefully valuable for the research community, including the importance of planning for this approach.

While the implementation presented relies on a vector store as an intuitive way for setting up the tool library, alternative RAG strategies and combinations thereof may be highly efficient and should be explored further. Also, the `tulip agent` architecture provides a basis for continuously extending the tool library with new tools created by learning from interactions, making it a sensible basis for open-ended applications, especially in robotics.

Tulip Agent    A Preprint# References

Tianle Li, Ge Zhang, Quy Duc Do, Xiang Yue, and Wenhu Chen. Long-context llms struggle with long in-context learning. *arXiv preprint arXiv:2404.02060*, 2024.

LangChain. Multi needle in a haystack, 2024. URL https://blog.langchain.dev/multi-needle-in-a-haystack/.

Stuart J Russell and Peter Norvig. *Artificial intelligence: a modern approach*. Pearson, 2016.

Stan Franklin and Art Graesser. Is it an agent, or just a program?: A taxonomy for autonomous agents. In *International workshop on agent theories, architectures, and languages*, pages 21–35. Springer, 1996.

Yingqiang Ge, Yujie Ren, Wenyue Hua, Shuyuan Xu, Juntao Tan, and Yongfeng Zhang. Llm as os (llmao), agents as apps: Envisioning aios, agents and the aios-agent ecosystem. *arXiv preprint arXiv:2312.03815*, 2023.

Jason Wei, Xuezhi Wang, Dale Schuurmans, Maarten Bosma, Fei Xia, Ed Chi, Quoc V Le, Denny Zhou, et al. Chain-of-thought prompting elicits reasoning in large language models. In *NeurIPS*, 2022.

Andrew Ng. Agentic design patterns. https://www.deeplearning.ai/the-batch/how-agents-can-improve-llm-performance/, 2024. Last accessed: 2024-05-29.

Shunyu Yao, Dian Yu, Jeffrey Zhao, Izhak Shafran, Tom Griffiths, Yuan Cao, and Karthik Narasimhan. Tree of thoughts: Deliberate problem solving with large language models. In *NeurIPS*, 2023.

Di Zhang, Jiatong Li, Xiaoshui Huang, Dongzhan Zhou, Yuqiang Li, and Wanli Ouyang. Accessing gpt-4 level mathematical olympiad solutions via monte carlo tree self-refine with llama-3 8b. *arXiv preprint arXiv:2406.07394*, 2024.

Yujia Qin, Shengding Hu, Yankai Lin, Weize Chen, Ning Ding, Ganqu Cui, Zheni Zeng, Yufei Huang, Chaojun Xiao, Chi Han, et al. Tool learning with foundation models. *arXiv preprint arXiv:2304.08354*, 2023a.

Timo Schick, Jane Dwivedi-Yu, Roberto Dessì, Roberta Raileanu, Maria Lomeli, Eric Hambro, Luke Zettlemoyer, Nicola Cancedda, and Thomas Scialom. Toolformer: Language models can teach themselves to use tools. In *NeurIPS*, 2023.

Pan Lu, Baolin Peng, Hao Cheng, Michel Galley, Kai-Wei Chang, Ying Nian Wu, Song-Chun Zhu, and Jianfeng Gao. Chameleon: Plug-and-play compositional reasoning with large language models. In *NeurIPS*, 2023.

Shijue Huang, Wanjun Zhong, Jianqiao Lu, Qi Zhu, Jiahui Gao, Weiwen Liu, Yutai Hou, Xingshan Zeng, Yasheng Wang, Lifeng Shang, et al. Planning, creation, usage: Benchmarking llms for comprehensive tool utilization in real-world complex scenarios. *arXiv preprint arXiv:2401.17167*, 2024.

Qiaoyu Tang, Ziliang Deng, Hongyu Lin, Xianpei Han, Qiao Liang, and Le Sun. Toolalpaca: Generalized tool learning for language models with 3000 simulated cases. *arXiv preprint arXiv:2306.05301*, 2023.

Shunyu Yao, Jeffrey Zhao, Dian Yu, Nan Du, Izhak Shafran, Karthik Narasimhan, and Yuan Cao. React: Synergizing reasoning and acting in language models. In *ICLR*, 2022.

Wei Chen and Zhiyuan Li. Octopus v2: On-device language model for super agent. *arXiv preprint arXiv:2404.01744*, 2024.

Wei Chen, Zhiyuan Li, and Mingyuan Ma. Octopus: On-device language model for function calling of software apis. *arXiv preprint arXiv:2404.01549*, 2024.

AI @ Meta Llama Team. The llama 3 herd of models. *arXiv preprint*, 2024.

Noah Shinn, Federico Cassano, Ashwin Gopinath, Karthik Narasimhan, and Shunyu Yao. Reflexion: Language agents with verbal reinforcement learning. In *NeurIPS*, 2023.

Yuchen Zhuang, Xiang Chen, Tong Yu, Saayan Mitra, Victor Bursztyn, Ryan A Rossi, Somdeb Sarkhel, and Chao Zhang. Toolchain*: Efficient action space navigation in large language models with a* search. *arXiv preprint arXiv:2310.13227*, 2023.

Zhaoyang Liu, Zeqiang Lai, Zhangwei Gao, Erfei Cui, Zhiheng Li, Xizhou Zhu, Lewei Lu, Qifeng Chen, Yu Qiao, Jifeng Dai, et al. Controlllm: Augment language models with tools by searching on graphs. *arXiv preprint arXiv:2310.17796*, 2023a.

Haotian Sun, Yuchen Zhuang, Lingkai Kong, Bo Dai, and Chao Zhang. Adaplanner: Adaptive planning from feedback with language models. In *NeurIPS*, 2023.
14

# Appendix

**Agent Prompts**

Prompts used, see `prompts.py` in the repository.

Listing 1: Baseline CoT prompt for the `NaiveToolAgent`, for use without a tool library.
```
1  You are a helpful agent who has access to an abundance of tools.
2  Always adhere to the following procedure:
3  1. Identify all individual steps mentioned in the user request.
4  2. Whenever possible use the tools available to fulfill the user request.
5  3. Respond to the user with the final result.
```

Listing 2: `CotTulipAgent` system prompt.
```
1  You are a helpful agent who has access to an abundance of tools.
2  Always adhere to the following procedure:
3  1. Break the user request down into atomic tasks.
4  2. Search your tool library for appropriate tools for these atomic tasks using the `search_tools` function. Provide
       generic task descriptions to ensure that you find generic tools.
5  3. Whenever possible use the tools found to solve the atomic tasks.
6  4. Respond to the user with the final result, never with an intermediate result.
```

Listing 3: Task decomposition prompt.
```
1  Considering the following task, what are the necessary steps you need to execute?
2  `{prompt}`
3  Return an ordered list of steps.
4  Return valid JSON and use the key `subtasks`.
```

Listing 4: Tool search prompt.
```
1  Search for suitable tools for each of the following tasks:
2  {tasks}
```

Listing 5: Prompt for solving a task with tools.

```
1  Now use the tools to fulfill the user request. Adhere exactly to the following steps:
2  {steps}
3  Execute the tool calls one at a time.
```

**Exemplary Tools**

Exemplary tools from the mathematics domain and the robotics domain.

Listing 6: Tool `add`.

```python
def add(a: float, b: float) -> float:
    """
    Add two numbers.

    :param a: The first number.
    :param b: The second number.
    :return: The sum of a and b.
    """
    return a + b
```

Listing 7: Tool `multiply`.

```python
def multiply(a: float, b: float) -> float:
    """
    Multiply two numbers.

    :param a: The first multiplicand.
    :param b: The second multiplicand.
    :return: The product of a and b.
    """
    return a * b
```

Listing 8: Tool `coefficient_of_variation`.

```python
def coefficient_of_variation(numbers: list[float]) -> float:
    """
    Calculate the coefficient of variation of a list of numbers.

    :param numbers: A list of numbers.
    :return: The coefficient of variation.
    """
    stdev = standard_deviation(numbers)
    mean = sum(numbers) / len(numbers)
    return stdev / mean
```

Listing 9: Tool `pour_into`.

```python
def pour_into(source_container_name: str, target_container_name: str) -> str:
    """
    You get a source container, pour it into a target container, and put it back on the table.

    :param source_container_name: The name of the container to pour from.
    :param target_container_name: The name of the container to pour into.
    :return: Result message.
    """
    success = SIMULATION.planActionSequence(
        (
            f"get {source_container_name} duration 8;"
            f"pour {source_container_name} {target_container_name};"
            f"put {source_container_name} table duration 7;"
            "pose default duration 4"
        ),
        ARG1,
    )
    if success:
        return f"You poured {source_container_name} into {target_container_name}."
    return f"You were not able to pour {source_container_name} into {target_container_name}."
```

**Tasks**

Listing 10: Excerpt of the ground truth showing task M.T.M.003.

```
1  [
2      ...,
3      {
4          "task": "Calculate 10th Fibonacci number and add it to the factorial of 10.",
5          "functions": [
6              "fibonacci_recursive",
7              "factorial",
8              "add"
9          ],
10         "name": "M.T.M.003",
11         "valid_solutions": [
12             3628855,
13             "3,628,855"
14         ]
15     },
16     ...
17 ]
```

**Costs**

Costs for the embedding and language models used[12] are summarized in Table 5.

Table 5: Inference and embedding costs.

|  | Costs per 1M input tokens | Costs per 1M output tokens |
| --- | --- | --- |
| `gpt-4-turbo-2024-04-09` | 10.00$ | 30.00$ |
| `gpt-3.5-turbo-0125` | 0.50$ | 1.50$ |
| `text-embedding-ada-002` | 0.10$ |  |
| `text-embedding-3-small` | 0.02$ |  |
| `text-embedding-3-large` | 0.13$ |  |

**Outtakes**

Exemplary logs for `NaiveToolAgent` and `CotTulipAgent` for the task "What is $45342 \cdot 23487 + 32478$?", and a CRUD example for `AutoTulipAgent`.

Listing 11: `NaiveToolAgent` logs for "What is $45342 \cdot 23487 + 32478$?".

```
1  INFO:root:Query: What is 45342 * 23487 + 32478?
2  INFO:base_agent:Usage for chatcmpl-94zcjnURFRF2nyTMlcVzhWJmfjdTZ in tokens: 3960 prompt and 19 completion.
3  INFO:base_agent:Function multiply returned `1064947554` for arguments {'a': 45342, 'b': 23487}.
4  INFO:base_agent:Usage for chatcmpl-94zcliQksrQsOMn72GamhLxiJRqN3 in tokens: 3990 prompt and 21 completion.
5  INFO:base_agent:Function add returned `1064980032` for arguments {'a': 1064947554, 'b': 32478}.
6  INFO:base_agent:Usage for chatcmpl-94zcyDN3oBC8rh5hzFNZkn1iXakLJ in tokens: 4022 prompt and 25 completion.
7  base_res='The result of \\(45342 \\times 23487 + 32478\\) is \\(1064980032\\).'
```

Listing 12: `CotTulipAgent` logs for "What is $45342 \cdot 23487 + 32478$?".

```
1  INFO:root:Query: What is 45342 * 23487 + 32478?
2  INFO:tulip_agent:Usage for chatcmpl-94zjNuHaG36iH6FtEuE4poOgRhTG4 in tokens: 182 prompt and 25 completion.
3  INFO:root:actions_response_message=ChatCompletionMessage(content='1. Multiply 45342 by 23487.\n2. Add the result of
       step 1 to 32478.', role='assistant', function_call=None, tool_calls=None)
4  INFO:tulip_agent:Usage for chatcmpl-94zjPPlK1ySBo6VZJLFMZhPyEONJK in tokens: 234 prompt and 13 completion.
5  INFO:root:Tool search for: {'action_descriptions': ['multiply two numbers', 'add two numbers']}
6  INFO:root:Functions for `multiply two numbers`: [...]
7  INFO:root:Functions for `add two numbers`: [...]
8  INFO:tulip_agent:Usage for chatcmpl-94zjRk9IrLzy5QKpfNEs5bsrXCGvp in tokens: 413 prompt and 19 completion.
9  INFO:tulip_agent:Function multiply returned `1064947554` for arguments {'a': 45342, 'b': 23487}.
10 INFO:tulip_agent:Usage for chatcmpl-94zjTXb58ZHJiNYDugqpwpaeffzzB in tokens: 443 prompt and 21 completion.
11 INFO:tulip_agent:Function add returned `1064980032` for arguments {'a': 1064947554, 'b': 32478}.
12 INFO:tulip_agent:Usage for chatcmpl-94zjVrSx0sLQN9ZqoI8f4z6JWOeVx in tokens: 475 prompt and 25 completion.
13 tulip_res='The result of \\(45342 \\times 23487 + 32478\\) is \\(1064980032\\).'
```

---

[12]https://openai.com/pricing, as of July 31, 2024

Listing 13: Log of a fully autonomous application of the `AutoTulipAgent`.

```
1  >> What is the square root of 23456789?
2  - Tool search for: {'action_descriptions': ['calculate square root']}
3  - Tools found: []
4  - Creating tool: {'task_description': 'calculate the square root of a number'}
5  - Made tool `calculate_square_root_module__calculate_square_root` available via the tool library.
6  - Function calculate_square_root_module__calculate_square_root returned `4843.220932396126` for arguments {'number':
       23456789}.
7  << The square root of 23,456,789 is approximately 4843.22.
8
9  >> Change the square root tool to correctly work for negative numbers.
10 - Updating tool: {'tool_name': 'calculate_square_root_module__calculate_square_root', 'instruction': 'Update the tool
       to handle negative numbers by returning complex numbers when the input is negative.'}
11 - Successfully generated code for the task `Edit the following Python code according to the instruction. Make sure to
       not change function names in the process. ...`
12 - Successfully updated `calculate_square_root_module__calculate_square_root`.
13 << The square root tool has been successfully updated to correctly handle negative numbers by returning complex numbers
       when the input is negative.
14
15 >> Calculate the square root of -200.
16 - Function calculate_square_root_module__calculate_square_root returned `14.142135623730951j` for arguments {'number
       ':-200}.
17 << The square root of -200 is approximately 14.14i.
18
19 >> Delete the square root tool.
20 - Deleting tool: {'tool_name': 'calculate_square_root_module__calculate_square_root'}
21 << The square root tool has been successfully deleted from the tool library.
```

**Tool Generation for the Benchmark**

Existing benchmarks for evaluating LLMs regarding tool use focus on the correctness of the tool calls. In contrast, we require a detailed analysis of the tool use, including precision and recall of tool calls including the parameters specified. A mathematics benchmark is well-suited for assessing the benefits of the `tulip agent` architecture, as complex tasks can easily be created that exceed the abilities of off-the-shelf LLMs.

To generate a large number of Python functions for mathematics tasks we used an LLM for code generation, specifically `gpt-4-0125-preview`. The LLM is initialized with the system prompt shown in Listing 14. All functions are generated with suitable documentation, including type hints and docstrings with parameter descriptions, see Listing 15, and Listing 16 reduces duplicate functions.

Note that the prompts include the following variables:

- `SUBFIELD`: 10 subfields of mathematics - algebra, analysis, calculus, number theory, geometry, topology, logic, set theory, probability theory, and statistics.
- `NUMBER_FUNCTIONS`: 5 new functions per iteration, to avoid model laziness.
- `KNOWN_FUNCTIONS`: Functions already generated, helps reduce duplicate functions.

Listing 14: System prompt for generating Python tools.

```
1  You are a very senior Python developer.
2  You are extremely efficient and return ONLY code.
```

Listing 15: Prompt for generating Python tools.

```
1  Please write NUMBER_FUNCTIONS Python functions for solving math tasks related to SUBFIELD.
2  You may include even trivial functions, such as addition and subtraction.
3  Adhere to the following rules:
4  1. Use sphinx documentation style without type documentation
5  2. Add meaningful and slightly verbose docstrings
6  3. Use python type hints
7  4. Return only valid code and avoid Markdown syntax for code blocks
8  5. Avoid adding examples to the docstring
```

Listing 16: Optional prompt for avoiding duplicates in Python tools.

```
1  Make sure to return unique functions and do not include the following ones: KNOWN_FUNCTIONS.
```

In total, we ran 25 iterations for all of the 10 subfields generating 5 functions each. After manual checks and deduplication, we ended up with a total of 605 tools. For the code see the `generate_eval_functions` module in the project repository's `eval` directory.